\documentclass[11pt,a4paper]{article}
\usepackage[hyperref]{acl2019}
\usepackage{times}
\usepackage{latexsym}
\usepackage{booktabs}
\usepackage{amsmath}
\usepackage{booktabs}  
\usepackage{amsmath} 
\usepackage{amssymb}   
\usepackage{mathtools}
\usepackage{tikz} 
\usetikzlibrary{shapes, arrows, positioning, fit, calc, backgrounds, shadows, arrows.meta}
\usepackage{listings}
\usepackage{xcolor}
\usepackage{microtype}

\usepackage{url}

\aclfinalcopy 

\title{Mitigating Spurious Correlations in NLI via LLM-Synthesized Counterfactuals and Dynamic Balanced Sampling}

\author{Christopher Román Jaimes \\
  University of Texas at Austin \\
  \texttt{romanjaimesc@utexas.edu}
}

\date{}

\begin{document}
\maketitle
\begin{abstract}
Natural Language Inference (NLI) models frequently rely on spurious correlations rather than semantic reasoning. Existing mitigation strategies often incur high annotation costs or trigger catastrophic forgetting during fine-tuning. We propose an automated, scalable pipeline to address these limitations. First, we introduce Log-Frequency LMI (LF-LMI) to accurately detect semantic artifacts. Second, we generate a high-quality synthetic contrast set via an LLM-synthesis pipeline with multi-judge verification. Finally, we introduce Dynamic Balanced Sampling, a training strategy that rotates the original data distribution to prevent forgetting. Our method improves consistency on a challenging benchmark from 63.5\% to 81.0\% while maintaining 88.4\% in-domain accuracy, significantly outperforming naive fine-tuning.
\end{abstract}

\section{Introduction}

Natural Language Inference (NLI) is a fundamental task in understanding textual entailment, requiring a model to determine whether a \textit{hypothesis} is true (\textsc{Entailment}), false (\textsc{Contradiction}), or unrelated (\textsc{Neutral}) given a \textit{premise}. Ideally, successful performance requires reasoning about the relationship between both sentences. However, prior research has demonstrated that standard NLI models often bypass this reasoning process, relying instead on superficial patterns or ``annotation artifacts'' present in the hypothesis alone \cite{gururangan2018, poliak2018}. 
For instance, models trained on the SNLI dataset \cite{bowman2015large} frequently learn that negation words predict \textsc{Contradiction}, achieving deceptively high accuracy without attending to the premise. This phenomenon forces us to rethink whether our datasets genuinely capture the core phenomenon we intend to model—NLI reasoning—or merely provide statistical shortcuts that facilitate prediction.

While rigorous evaluation frameworks like contrast sets \cite{gardner2020} and symmetric datasets \cite{schuster2019} exist, they rely on expensive human annotation. Conversely, automated data augmentation \cite{wu2021polyjuice} offers scalability but often triggers catastrophic forgetting during fine-tuning. To address these limitations, we propose an automated pipeline with three key contributions: 

\begin{enumerate} 

    \item \textbf{Artifact Detection:} We introduce \textbf{LF-LMI}, a modified LMI metric that dampens frequency noise to precisely identify semantic artifacts. 
    
    \item \textbf{Automated Synthesis:} We develop an LLM-driven pipeline with multi-judge verification to generate high-quality contrast sets, combining manual rigor with automated scalability. 
    
    \item \textbf{Robust Training:} We propose \textbf{Dynamic Balanced Sampling}, a strategy that rotates the original data distribution to prevent forgetting. 

\end{enumerate} 

Our method improves consistency on a challenging benchmark from $63.5\%$ to $81.0\%$ while maintaining $88.4\%$ in-domain accuracy.

\section{Related Work}

\paragraph{Spurious Correlations in NLI.}
The reliance of NLI models on superficial patterns is a well-established challenge. Previous works \cite{gururangan2018, poliak2018} revealed that crowd-workers often adopt heuristics---such as negating the premise to generate contradictions---creating ``annotation artifacts'' that allow models to shortcut reasoning. \newcite{schuster2019} further analyzed how these biases propagate to model predictions. Our work leverages these insights, using our proposed LF-LMI metric to systematically detect these semantic artifacts before they influence training.

\paragraph{Data-Centric Mitigation Strategies.}
To measure and improve robustness, \newcite{gardner2020} introduced Contrast Sets, manually perturbing test examples to flip the label and challenge local decision boundaries. While highly effective, manual curation is costly and unscalable. Automated augmentation methods like Polyjuice \cite{wu2021polyjuice} offer scalability but often lack the strict logical consistency required for rigorous NLI debiasing. Our approach bridges this gap by automating the generation of high-quality contrast sets using LLMs with a multi-judge verification protocol.

\paragraph{Catastrophic Forgetting.}
Mitigating biases often requires fine-tuning on counterfactual or ``hard'' data distributions. However, sequential training on these shifts frequently leads to catastrophic forgetting, where the model degrades on the original distribution \cite{luo2023empirical}. Unlike naive fine-tuning approaches, our Dynamic Balanced Sampling strategy is explicitly designed to maintain in-domain performance while improving out-of-domain robustness.

\section{Diagnostic Analysis}

\subsection{Evidence of Spurious Correlation}
\label{sec:evidence}
When an NLI model is trained, we ideally expect it should predict the relationship between a premise and a hypothesis by reasoning over both sentences. To verify if this holds true, we evaluated the behavior of a standard classifier when trained using the hypothesis alone.

We trained a baseline model ELECTRA-Small, on the SNLI training set using the standard input (premise + hypothesis), achieving an accuracy of $89.4\%$, significantly outperforming the majority class baseline of $33.3\%$ (given the three balanced classes). However, when we trained the same architecture in a hypothesis-only setting \cite{poliak2018}, i.e., disregarding the input premise, the model achieved an accuracy of $61.6\%$.

This result leads to two critical conclusions. First, the hypothesis-only performance is nearly double that of the baseline, confirming the existence of strong spurious correlations between hypothesis patterns and labels that allow the model to shortcut the reasoning process. Second, following \citet{poliak2018}, we argue that the baseline for NLI should not be the majority class, but rather the hypothesis-only performance. Any performance above $33.3\%$ but below $61.6\%$ likely relies on spurious correlations rather than true NLI reasoning. Consequently, standard evaluations drastically underestimate the severity of the problem.

\subsection{Detecting Artifacts via LF-LMI}

\begin{table*}[t] 
\centering
\small
\begin{tabular}{cc}
    \begin{tabular}{lrrr}
        \toprule
        \textbf{Bigram} & \textbf{LF-LMI} & \textbf{Freq} & \textbf{$P(l|w)$} \\
        \midrule
        \textit{nobody is}   & 8.26 & 1784 & 1.00 \\
        \textit{is sleeping} & 7.75 & 3737 & 0.86 \\
        \textit{at home}     & 7.74 & 1848 & 0.93 \\
        \textit{are sleeping}& 7.57 & 1513 & 0.94 \\
        \textit{watching tv} & 7.24 & 1119 & 0.94 \\
        \textit{sleeping in} & 6.99 & 1297 & 0.89 \\
        \textit{a cat}       & 6.92 & 1718 & 0.85 \\
        \textit{no one}      & 6.78 &  733 & 0.93 \\
        \textit{swimming in} & 6.76 & 1844 & 0.83 \\
        \textit{in bed}      & 6.60 &  593 & 0.94 \\
        \textit{are no}      & 6.58 &  544 & 0.95 \\
        \textit{is asleep}   & 6.51 & 1070 & 0.86 \\
        \textit{cat is}      & 6.49 &  726 & 0.90 \\
        \textit{nobody has}  & 6.45 &  340 & 1.00 \\
        \textit{is no}       & 6.44 &  524 & 0.94 \\
        \bottomrule
    \end{tabular} 
    & 
    \begin{tabular}{lrrr}
        \toprule
        \textbf{Bigram} & \textbf{LMI} & \textbf{Freq} & \textbf{$P(l|w)$} \\
        \midrule
        \textit{in the}      & 3269.00 & 33127 & 0.42 \\
        \textit{is sleeping} & 3097.42 &  3737 & 0.86 \\
        \textit{on the}      & 2843.76 & 22755 & 0.44 \\
        \textit{is sitting}  & 2764.88 &  8039 & 0.59 \\
        \textit{in a}        & 2177.85 & 34691 & 0.39 \\
        \textit{nobody is}   & 1961.50 &  1784 & 1.00 \\
        \textit{sitting on}  & 1862.99 &  5774 & 0.58 \\
        \textit{at home}     & 1792.07 &  1848 & 0.93 \\
        \textit{the man}     & 1744.56 & 31938 & 0.38 \\
        \textit{are sitting} & 1606.45 &  5278 & 0.57 \\
        \textit{are sleeping}& 1479.21 &  1513 & 0.94 \\
        \textit{man is}      & 1472.41 & 56721 & 0.36 \\
        \textit{sitting in}  & 1438.38 &  3369 & 0.64 \\
        \textit{swimming in} & 1413.22 &  1844 & 0.83 \\
        \textit{a cat}       & 1393.60 &  1718 & 0.85 \\
        \bottomrule
    \end{tabular} \\
    (a) LF-LMI (Ours) & (b) Standard LMI
\end{tabular}
\caption{Top-15 spurious bigrams for the \textsc{Contradiction} label. (a) Shows bigrams retrieved by our Log-Frequency LMI, which successfully surfaces negation artifacts. (b) Shows bigrams from the standard LMI, which is dominated by high-frequency stopwords with low predictive probability $P(l|w)$.}
\label{tab:lmi_comparison}
\end{table*}

While the hypothesis-only baseline confirms the presence of the unexpected correlation, it does not identify the specific linguistic features driving it. These features, known as annotation artifacts, are well documented in the literature. For example, \citet{schuster2019} found that negative bigrams such as \textit{did not} correlate strongly with the \textsc{Contradiction} label in the FEVER dataset. Similarly, \citet{gururangan2018} demonstrated that in SNLI, negation words like \textit{nobody} and \textit{no} are strong predictors of \textsc{Contradiction}, while generalizations like \textit{outdoors} or \textit{animal} signal \textsc{Entailment}.

To systematically detect the patterns that enable these spurious correlations, we rely on Local Mutual Information (LMI) \cite{evert2005}. LMI helps identify words that disproportionately co-occur with a specific label by weighting their conditional probability with their joint frequency. However, standard LMI is biased towards high-frequency terms. We propose a modified version, \textbf{Log-Frequency LMI (LF-LMI)}, defined as: 
\begin{equation}
\mathrm{LF\text{-}LMI}(w, l) = 
\!\mathcal{F}_{log}(w, l) \cdot 
\log\!\left(\frac{P(l \mid w)}{P(l)}\right)
\end{equation}

Where $\mathcal{F}_{log}(w, l) = \log(count(w, l))$ denotes the log-transformed frequency of the n-gram $w$ co-occurring with label $l$ and the probability terms $P(l|w)$ and $P(l)$ are estimated via maximum likelihood from the training set counts. By applying a logarithmic scale to the frequency term, LF-LMI reduces the dominance of extremely frequent but uninformative n-grams (such as stop words), allowing semantic artifacts to emerge as top features. 

To validate this approach, we compared the top-15 bigrams retrieved by standard LMI against our LF-LMI using the SNLI training data for the \textsc{Contradiction} label. The results show three main improvements when using the log-frequency adjustment:

\begin{itemize}
    \item \textbf{Filtering noise:} The standard LMI placed \textit{in the} as the number one which is clearly an uninformative stopword bi-gram and it is ranked highly for being extremely common. On the other hand, LF-LMI identified \textit{nobody is}  as the top artifact, which fits perfectly with well-established findings regarding negation bias in NLI \citet{poliak2018}.

    \item \textbf{Better Signal Quality:} As you can see in Table~\ref{tab:lmi_comparison}, the candidates from the standard LMI are highly frequent but basically uninformative phrases like \textit{in a} or \textit{on the}. In fact, many of these have a conditional probability $P(l|w)$ below 0.5, meaning they do not offer much predictive power at all. Our LF-LMI successfully attenuates frequency terms, every single one of the top-15 bigrams it retrieved have a conditional probability higher than 0.85.

    \item \textbf{Catching the Semantic Artifacts:} When we looked at the top-15 rank, the standard LMI only caught one explicit negation bigram (\textit{nobody is}). In contrast, LF-LMI found four more (\textit{no one}, \textit{are no}, \textit{nobody has}, \textit{is no}). This demonstrates that our metric is much more sensitive to the specific linguistic patterns that models tend to exploit as shortcuts.

\end{itemize}

We observed similar improvements for \textsc{Entailment} and \textsc{Neutral} labels; full tables are provided in Appendix~\ref{app:lmi_tables}.

\subsection{Anatomy of Dataset Artifacts}

Using some artifacts retrieved by LF-LMI, we can categorize the spurious correlations into three distinct linguistic heuristics used by annotators, which explains the reason behind each artifact.

\paragraph{Negation as a Proxy for \textsc{Contradiction}.}As demonstrated in the LF-LMI calculation step, negation markers such as \textit{nobody is}, \textit{no one}, and \textit{is no} are strongly correlated with the \textsc{Contradiction} label. For instance, the bigrams \textit{nobody is} and \textit{nobody has} exhibit a $P(l|w)$ of exactly $1.0$, meaning that in every instance these tokens appear, the label is \textsc{Contradiction}. This bias likely stems from the cognitive ease of generating a contradictory hypothesis by simply negating the premise (e.g., \textit{A boy in a large pool} $\rightarrow$ \textit{Nobody is in a pool}).\footnote{In visual-grounded NLI datasets derived from image captions (e.g., SNLI), reasoning typically operates under a local \textit{Closed World Assumption}: the premise is treated as an exhaustive description of all salient entities and actions in the scene. Consequently, any salient object or action not explicitly mentioned in the premise is inferred to be absent from the image.}

\paragraph{Generalization as a Proxy for \textsc{Entailment}.}A similar analysis for \textsc{Entailment} reveals that bigrams representing high-level generalizations, such as \textit{a human}, \textit{an animal}, or \textit{is outside}, are strong predictors of the label. This reflects a strategy where annotators generate true hypotheses by replacing specific nouns with hypernyms and specific locations with vague prepositions. For example, in the pair \textit{A dog runs in a field} $\rightarrow$ \textit{An animal is outside}, the specific entity \textit{dog} is generalized to \textit{animal}, and the specific setting \textit{field} is generalized to \textit{outside}.

\paragraph{Inference as a Proxy for \textsc{Neutral}.}Finally, bigrams implying future intent or internal states, such as \textit{going to}, \textit{waiting for}, \textit{practicing for}, or \textit{is winning}, hold strong predictive power for the \textsc{Neutral} label. This correlates with annotators writing plausible but unverifiable hypotheses based on the visual scene. For instance, consider the pair \textit{Guy on his dirt bike doing tricks} $\rightarrow$ \textit{The man is practicing for BMX}. While the hypothesis is plausible, this hypothesis is not logically entailed because the premise does not specify the subject's intent (he could be doing tricks for fun rather than competition).

\section{Proposed Method}

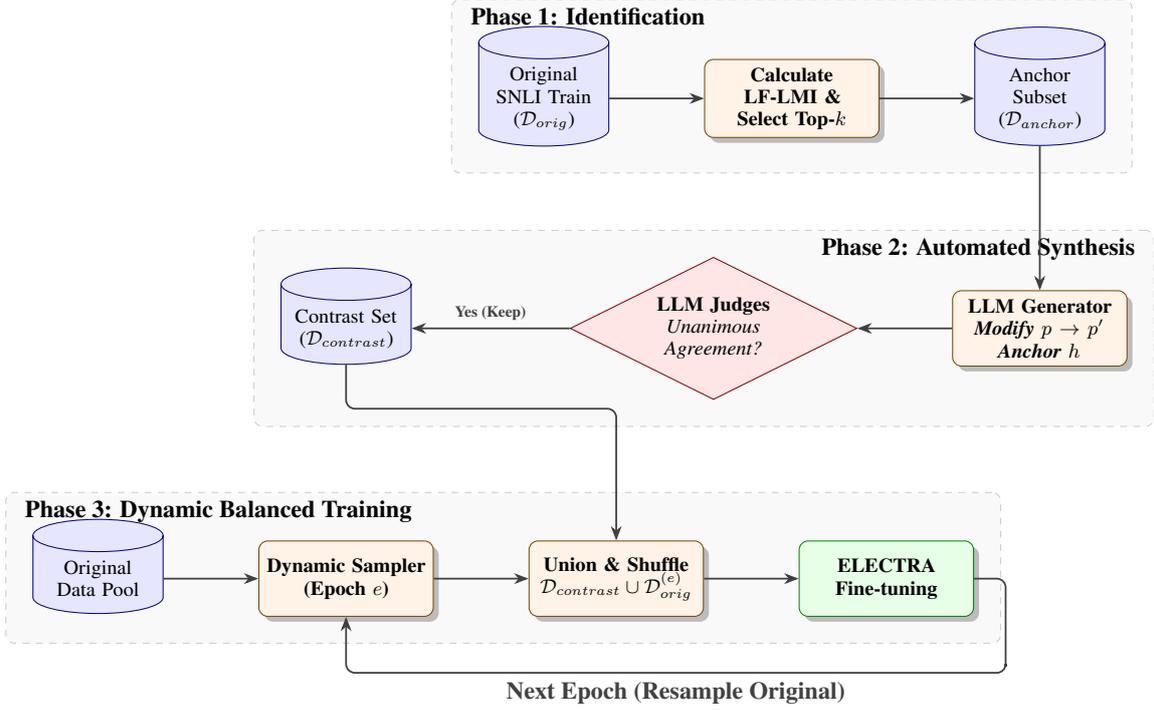
\begin{figure*}[t]
\centering
\resizebox{0.95\textwidth}{!}{%
\begin{tikzpicture}[
    node distance=1.8cm and 1.5cm,
    dataset/.style={cylinder, shape border rotate=90, draw=blue!50!black, fill=blue!10, aspect=0.25, text width=1.8cm, align=center, minimum height=1.5cm, font=\footnotesize},
    process/.style={rectangle, draw=orange!50!black, fill=orange!10, rounded corners, text width=2.5cm, align=center, minimum height=1.2cm, drop shadow, font=\footnotesize\bfseries},
    decision/.style={diamond, draw=red!50!black, fill=red!10, aspect=2, text width=1.8cm, align=center, font=\footnotesize},
    arrow/.style={-Stealth, thick, color=gray!50!black, rounded corners},
    groupbox/.style={draw=gray!40, dashed, inner sep=12pt, fill=gray!5, rounded corners}
    ]

    \node [dataset] (snli) {Original SNLI Train ($\mathcal{D}_{orig}$)};
    \node [process, right=of snli] (lmi) {Calculate \textbf{LF-LMI} \& Select Top-$k$};
    \node [dataset, right=of lmi] (anchor) {Anchor Subset ($\mathcal{D}_{anchor}$)};

    \node [process, below=of anchor, yshift=-0.5cm] (gen) {\textbf{LLM Generator}\\ \textit{Modify $p \to p'$}\\ \textit{Anchor $h$}};
    \node [decision, left=of gen] (judge) {\textbf{LLM Judges}\\ \textit{Unanimous Agreement?}};
    \node [dataset, left=of judge, xshift=-1cm] (contrast) {Contrast Set ($\mathcal{D}_{contrast}$)};
    
    \node [process, below=of contrast, yshift=-1.0cm] (sampler) {\textbf{Dynamic Sampler}\\ (Epoch $e$)};
    \node [dataset, left=of sampler] (orig_pool) {Original Data Pool};
    \node [process, right=of sampler] (mix) {Union \& Shuffle\\ $\mathcal{D}_{contrast} \cup \mathcal{D}_{orig}^{(e)}$};
    \node [process, right=of mix, fill=green!10, draw=green!50!black] (model) {\textbf{ELECTRA}\\ Fine-tuning};

    \draw [arrow] (snli) -- (lmi);
    \draw [arrow] (lmi) -- (anchor);
    
    \draw [arrow] (anchor) -- (gen);
    \draw [arrow] (gen) -- (judge);
    
    \draw [arrow] (judge) -- node[above, font=\scriptsize] {\textbf{Yes (Keep)}} (contrast);
    
    \draw [arrow] (contrast.south) -- ++(0,-0.7) -| (mix.north);
    \draw [arrow] (orig_pool) -- (sampler);
    \draw [arrow] (sampler) -- (mix);
    \draw [arrow] (mix) -- (model);
    
    \draw [arrow] (model.east) -- ++(0.5,0) |- ++(0,-1.5) -| node[pos=0.25, below, font=\bfseries] {Next Epoch (Resample Original)} (sampler.south);

    \begin{scope}[on background layer]
        \node [groupbox, fit=(snli) (lmi) (anchor), label={[anchor=north west, xshift=5pt]north west:\textbf{Phase 1: Identification}}] (box1) {};
        \node [groupbox, fit=(gen) (judge) (contrast), label={[anchor=north east, xshift=-5pt]north east:\textbf{Phase 2: Automated Synthesis}}] (box2) {};
        \node [groupbox, fit=(orig_pool) (sampler) (mix) (model), label={[anchor=north west, xshift=5pt]north west:\textbf{Phase 3: Dynamic Balanced Training}}] (box3) {};
    \end{scope}

\end{tikzpicture}
}
\caption{\textbf{Overview of our Automated Debiasing Pipeline.} (1) Artifacts are identified via LF-LMI. (2) LLMs synthesize counterfactuals, filtered by strict consensus. (3) The model is trained using \textit{Dynamic Balanced Sampling}: the static contrast set is mixed with a rotating random subset of the original data in each epoch to prevent catastrophic forgetting.}
\label{fig:methodology}
\end{figure*}

To address the spurious correlations identified in Section 3, we propose an end-to-end framework. As illustrated in Figure~\ref{fig:methodology}, our pipeline operates in three phases. Having established the \textbf{Identification} of artifacts via LF-LMI (Phase 1), this section details the subsequent remediation steps: the \textbf{Synthesis} of a hard contrast set via LLMs (Phase 2) and the \textbf{Dynamic Balanced Training} strategy (Phase 3) designed to mitigate forgetting.

\subsection{Generation of LLM-Synthesized Counterfactuals}
Motivated by our analysis, we developed a method to systematically reduce spurious correlations by creating a dataset of LLM-generated counterfactuals.

The process begins by gathering the top-$k$ bigrams with the highest LF-LMI scores from the training data. For each identified artifact, we sample $m$ instances, creating a focused candidate pool that directly targets the original dataset weaknesses. To transform these candidates into a robust dataset, our approach synthesizes two key methodological frameworks:

\begin{itemize}
    \item \textbf{Local Decision Boundaries:} Following the principles of \citet{gardner2020}, we generate contrast sets to challenge the decision boundary of the model. For each original observation $(p, h, l)$, where $p$ is the premise, $h$ is the hypothesis, and $l$ is the label, we synthesize a counterfactual observation $(p', h, l')$ where the label is flipped ($l \neq l'$). Unlike traditional approaches that modify the hypothesis, we perturb the premise ($p \to p'$). This forces the model to re-evaluate the logical relationship based on a minimally altered visual context. \textit{Target Labels:} We map \textsc{Entailment} $\leftrightarrow$ \textsc{Contradiction} to maximize the logical contrast. For \textsc{Neutral}, we split the targets 50/50 between the other two classes to mitigate introducing a new class imbalance bias.
    \item \textbf{Debiasing by Anchoring the Hypothesis:} Crucially, we keep the hypothesis $h$ fixed across the pair. Inspired by the symmetric datasets of \citet{schuster2019}, this strategy ensures that the artifact (contained in $h$) is no longer predictive. By construction, the problematic n-gram is now associated with two conflicting labels, effectively neutralizing its spurious correlation ($P(l|w) \approx 0.5$). 
\end{itemize}

\begin{table*}[t]
    \centering
    \small 
    \renewcommand{\arraystretch}{1.1} 
    \setlength{\tabcolsep}{4pt}
    
    \begin{tabular}{@{}c p{0.27\linewidth} p{0.27\linewidth} p{0.22\linewidth}@{}}
        \toprule
        \textbf{Label Shift} & \textbf{Original Premise} ($p$) & \textbf{Synthesized Counterfactual} ($p'$) & \textbf{Fixed Hypothesis} ($h$) \\
        \midrule
        
        \begin{tabular}{@{}c@{}}
            \textsc{Contradiction} \\
            $\downarrow$ \\
            \textsc{Entailment}
        \end{tabular} & 
        A white dog with long hair jumps to catch a red and green toy. & 
        A white dog with long hair \textbf{swims underwater} to catch a red and green toy. & 
        \textit{A white dog with long hair is swimming underwater.} \\
        \midrule
        
        \begin{tabular}{@{}c@{}}
            \textsc{Entailment} \\
            $\downarrow$ \\
            \textsc{Contradiction}
        \end{tabular} & 
        A boy in a red shirt and a boy in a yellow shirt are jumping on a trampoline outside. & 
        A boy in a red shirt and a boy in a yellow shirt are jumping on a trampoline \textbf{inside}. & 
        \textit{The boys are outside.} \\
        \midrule
        
        \begin{tabular}{@{}c@{}}
            \textsc{Neutral} \\
            $\downarrow$ \\
            \textsc{Contradiction}
        \end{tabular} & 
        A girl standing by a decorated structural beam poses for a picture. & 
        A girl standing by a decorated structural beam poses for a picture \textbf{alone}. & 
        \textit{A girl poses for a picture with her sister.} \\
        \midrule
        
        \begin{tabular}{@{}c@{}}
            \textsc{Neutral} \\
            $\downarrow$ \\
            \textsc{Entailment}
        \end{tabular} & 
        A man standing in front of a class of Asian students holding a picture of Santa Claus. & 
        A \textbf{tall} man standing in front of a class of Asian students holding a picture of Santa Claus. & 
        \textit{A tall human standing.} \\
        \bottomrule
    \end{tabular}
    \caption{Examples of generated counterfactuals illustrating label transitions. The Fixed Hypothesis remains constant to anchor the artifact. The LLM minimally modifies the Original Premise (changes highlighted in bold) to flip the logical relationship (Label Shift).}
    \label{tab:generated_examples}
\end{table*}

We automated this process using a multi-LLM pipeline. The \textit{Synthesis} phase employs an LLM prompted to act as an expert annotator (full prompts are provided in Appendix~\ref{app:prompts}), instructed to perform minimal edits to the premise by modifying only the subject, action, or setting to satisfy the new target label. Table~\ref{tab:generated_examples} provides representative examples of these synthesized counterfactuals, illustrating how minimal lexical changes in the premise successfully flip the logical relationship across all label transitions. We employed a multi-judge LLM system to verify perturbation, naturalness, and accuracy. To prioritize precision, we enforced a strict consensus filter, discarding any non-unanimous samples. Finally, we performed manual validations on the new validation set to verify the integrity of the automated system. Upon completion of this pipeline, we formally denote the set of synthesized counterfactuals as $\mathcal{D}_{synth}$ and the set of their corresponding original observations as $\mathcal{D}_{anchor}$. To enable the evaluation of contrast consistency, the final contrast set is defined as the union $\mathcal{D}_{contrast} \coloneq \mathcal{D}_{anchor} \cup \mathcal{D}_{synth}$, yielding a potential total of $2 \cdot k \cdot m$ paired observations, paired observations, excluding those rejected by the LLM-as-a-Judge protocol.

\subsection{Dynamic Balanced Sampling}
With the curated contrast set established, our goal is to refine the model to handle these challenging examples. 

A naive approach would be to fine-tune the model exclusively on this new synthetic dataset. However, this strategy poses a significant risk of \textit{catastrophic forgetting}, a phenomenon where neural networks abruptly lose the knowledge acquired from previous data distributions when trained sequentially on new tasks or datasets. While this issue is prominently documented \cite{luo2023empirical}, our results confirm that a similar phenomenon significantly impacts smaller discriminative architectures like ELECTRA.

To mitigate this risk, we propose a \textbf{Dynamic Balanced Sampling strategy}. Instead of fine-tuning on a static dataset, we construct a dynamic training set for each epoch $e$. This set consists of the union of our fixed synthetic contrast set $\mathcal{D}_{contrast}$ and a random subset of the original training data $\mathcal{D}_{orig}(e)$, sampled at the beginning of each epoch such that $|\mathcal{D}_{orig}(e)| = |\mathcal{D}_{contrast}|$, then the contrast set is $\mathcal{D}_{mix}(e) \coloneq \mathcal{D}_{contrast} \cup {D}_{orig}(e)$. This approach ensures a 1:1 balance between ``hard'' artifact-free examples and ``standard'' in-domain examples. Our dynamic sampling strategy improves robustness while preserving original accuracy.

\section{Experimental Setup}
\paragraph{Dataset.}
We evaluate our approach on the \textbf{Stanford Natural Language Inference (SNLI)} corpus \cite{bowman2015large}. This dataset consists of premise-hypothesis pairs derived from image captions in the \textbf{Flickr30k} dataset, grounding the logical reasoning in visual scenes. The dataset is divided into two primary splits:
\begin{itemize}
    \item \textbf{Training data} $\mathcal{D}_{orig}^{train}$: $\approx$550k premise-hypothesis pairs.
    \item \textbf{Validation set} $\mathcal{D}_{orig}^{val}$: $\approx$10k premise-hypothesis pairs.
\end{itemize}
\noindent These standard splits served as the training and evaluation basis for the diagnostic models (Baseline and Hypothesis-Only) discussed in Section~\ref{sec:evidence}.
On the other hand, for our robust fine-tuning strategy, we construct dynamic datasets for each epoch $e \in \{1,2,3\}$:
\begin{itemize}
    \item \textbf{Training data} $\mathcal{D}_{mix}^{train}(e)$: Consists of 58,524 pairs total. 14,631 are a balanced random sample of the top-20 LF-LMI bigrams from $\mathcal{D}_{orig}^{train}$, 14,631 are their corresponding synthesized contrast pairs, and the remaining 29,262 are a random sample from $\mathcal{D}_{orig}^{train}$.
    \item \textbf{Validation set} $\mathcal{D}_{contrast}^{val}$: Consists of 936 pairs. 468 are a balanced random sample of the top-10 LF-LMI bigrams from $\mathcal{D}_{orig}^{val}$, and 468 are their corresponding contrast pairs. Unlike $\mathcal{D}_{mix}^{train}(e)$, this set remains static across epochs to ensure a consistent evaluation benchmark, as it excludes the dynamic sampling of the original data.
\end{itemize}
Both $\mathcal{D}_{mix}^{train}(e)$ and $\mathcal{D}_{contrast}^{val}$ are constructed following the pipeline detailed in Section 3.

\paragraph{Model Architecture.}
For all experiments, we employed the \textbf{ELECTRA-Small} discriminator \cite{clark2020electra}. ELECTRA is well-known for being as competitive as BERT while being significantly more computationally efficient. We prioritize ELECTRA-Small to demonstrate the efficiency of our data-centric pipeline under constrained compute environments, setting a baseline for future scalability studies.

\paragraph{Implementation Details.}
Models were implemented using the summarized configuration in Table \ref{tab:hyperparameters}. The critical modification in our proposed method is the reduction of the learning rate to $1\text{e-}6$ during the fine-tuning phase to preserve the pre-trained knowledge. For the automated data generation pipeline, we employed \texttt{gpt-5-mini} for counterfactual synthesis and a heterogeneous judge panel (\texttt{gpt-5-mini}, \texttt{gemini-2.5-flash}) for consensus-based validation.

\begin{table}[h]
    \centering
    \small
    \renewcommand{\arraystretch}{1.2}
    \begin{tabular}{lc}
        \toprule
        \textbf{Hyperparameter} & \textbf{Value} \\
        \midrule
        Batch Size & 128 \\
        Epochs & 3 \\
        Optimizer & AdamW \\
        Scheduler & Linear \\
        Adam $\beta_1, \beta_2$ & 0.9, 0.999 \\
        Adam $\epsilon$ & $1\text{e-}8$ \\
        Max Sequence Length & 128 \\
        Learning Rate (Pre-train) & $5\text{e-}5$ \\
        Learning Rate (Fine-tune) & $1\text{e-}6$ \\
        \bottomrule
    \end{tabular}
    \caption{Hyperparameter configuration. Common settings are listed above, while the learning rate was adjusted specifically for the robust fine-tuning phase.}
    \label{tab:hyperparameters}
\end{table}

\paragraph{Evaluation Metrics.}
For our robust evaluation on the contrast sets, standard accuracy is insufficient as it does not capture the model's stability across perturbations. Following \citet{gardner2020}, we adopt the \textbf{Consistency Score} as our primary metric. Consistency is defined as the percentage of paired examples (original observation $(p, h, l)$ and its counterfactual $(p', h, l')$) where the model correctly predicts the label for \textit{both} instances. This metric is strictly more rigorous than accuracy, as it penalizes models that rely on spurious correlations to solve one instance while failing on its minimal perturbation.

\section{Empirical Results}

\subsection{Validation of the Synthetic Benchmark}

Before evaluating our proposed method, we first analyze the properties of the constructed synthetic validation set $\mathcal{D}_{contrast}^{val}$ to ensure it effectively challenges the model's reliance on artifacts.

\paragraph{Neutralizing Artifacts.}
A key objective of our generation pipeline was to break the statistical correlation between specific n-grams and labels. As shown in Table~\ref{tab:prob_shift}, the conditional probability $P(l|w)$ for top artifact bigrams, which was near $1.0$ in the original training data, has dropped to exactly $0.5$ in the validation contrast set. This confirms that for every instance where an artifact like \textit{nobody is} appears, there is now an equal probability of it belonging to the \textsc{Entailment} or \textsc{Contradiction} class, effectively removing the artifact's predictive power.

\begin{table}[h]
    \centering
    \small
    \renewcommand{\arraystretch}{1.2}
    \begin{tabular}{lcc}
        \toprule
        \textbf{Bigram} & \textbf{Original} $P(l|w)$ & \textbf{Contrast} $P(l|w)$ \\
        \midrule
        \textit{nobody is} & 1.00 & 0.50 \\
        \textit{nobody has} & 1.00 & 0.50 \\
        \textit{no one} & 0.93 & 0.50 \\
        \textit{is sleeping} & 0.86 & 0.50 \\
        \bottomrule
    \end{tabular}
\caption{Shift in conditional probabilities for the top artifacts identified by LF-LMI in the \textsc{Contradiction} class. In the original training set, these n-grams were strong predictors of the label. In our contrast set, their probability drops to chance level ($0.50$), effectively neutralizing the spurious correlation.}
    \label{tab:prob_shift}
\end{table}

\paragraph{Class Distribution Shift.}
It is important to note that the construction of $\mathcal{D}_{contrast}^{val}$ alters the class distribution compared to the original balanced SNLI set. In our generation protocol, original \textsc{Neutral} examples are mapped $50\%$ to \textsc{Entailment} and $50\%$ to \textsc{Contradiction}, while original \textsc{Entailment} and \textsc{Contradiction} are swapped. Consequently, by construction in the final combined set (Original + Counterfactuals), the \textsc{Neutral} class represents only $1/6$ ($\approx 16.6\%$) of the data, while \textsc{Entailment} and \textsc{Contradiction} each account for $\approx 41.7\%$ which is the new majority class baseline.

\begin{table*}[t]
    \centering
    \small
    \renewcommand{\arraystretch}{1.3}
    \setlength{\tabcolsep}{8pt}
    \begin{tabular}{lccc}
        \toprule
        \textbf{Model Strategy} & \textbf{Original Acc} ($\mathcal{D}_{orig}^{val}$) & \textbf{Contrast Acc} ($\mathcal{D}_{contrast}^{val}$) & \textbf{Consistency Score} ($\mathcal{D}_{contrast}^{val}$) \\
        \midrule
        Baseline (Pre-trained) & 89.4\% & 81.6\% & 63.5\% \\
        Naive Finetuning & 84.1\% \textcolor{red}{\scriptsize($\Delta$-5.3\%)} & \textbf{90.4\%} \textcolor{blue}{\scriptsize($\Delta$+8.8\%)} & \textbf{81.0\%} \textcolor{blue}{\scriptsize($\Delta$+17.5\%)} \\
        \textbf{Dynamic Balanced} & \textbf{88.4\%} \textcolor{red}{\scriptsize($\Delta$-1.0\%)} & 90.3\% \textcolor{blue}{\scriptsize($\Delta$+8.7\%)} & \textbf{81.0\%} \textcolor{blue}{\scriptsize($\Delta$+17.5\%)} \\
        \bottomrule
    \end{tabular}
    \caption{Performance comparison across training strategies. $\Delta$ values indicate the absolute percentage point change relative to the Baseline. Note that while Naive Finetuning achieves high contrast performance, it suffers a significant drop in original accuracy (catastrophic forgetting). Our \textbf{Dynamic Balanced} strategy achieves the same consistency gains while preserving in-domain performance.}
    \label{tab:main_results}
\end{table*}

\paragraph{The Gap Between Accuracy and Consistency.}
When evaluating the Baseline model on this new set, we observe an accuracy of $81.6\%$. While this represents a drop from the original $89.4\%$, it might still imply decent performance. However, this metric is misleading. The \textbf{Consistency Score}—which requires the model to correctly predict both the original and the counterfactual example—is significantly lower at $63.5\%$. This large gap reveals that the model is often ``right for the wrong reasons,'' correctly classifying the easy original example but failing its counterfactual pair. Furthermore, the Hypothesis-Only model collapses to  $49.3\%$, hovering near the majority class baseline of $41.7\%$ and yielding a consistency score of exactly $0.0\%$. This absolute zero consistency is expected by construction: since the hypothesis remains constant across pairs while the label flips, a model relying solely on hypothesis artifacts cannot possibly predict both instances correctly, confirming that $\mathcal{D}_{contrast}^{val}$ successfully neutralizes hypothesis-only cues.

Collectively, these results demonstrate that our contrast dataset successfully eliminates the spurious correlations between hypothesis artifacts and labels, presenting a significantly more challenging benchmark that exposes the reasoning fragility of standard models.

\subsection{Main Results: Dynamic Sampling Prevents Forgetting}

Starting from the pre-trained ELECTRA baseline, we evaluated two distinct fine-tuning strategies. Table~\ref{tab:main_results} presents the performance comparison, where we observe two key trends:

\textbf{Naive Finetuning:} Fine-tuning exclusively on the artifact-free dataset yields high performance on the contrast set ($90.4\%$ Accuracy, $81.0\%$ Consistency), proving that the model can learn to ignore artifacts. However, this comes at a steep cost: accuracy on the original validation set drops by $5.3\%$ ($89.4\% \to 84.1\%$). This confirms the occurrence of \textit{catastrophic forgetting}, where the model overfits to the hard distribution and loses general NLI capabilities.

\textbf{Dynamic Balanced Sampling (Ours):} Our proposed strategy effectively solves this trade-off. By rotating a balanced sample of the original data in each epoch, the model retains its original performance ($88.4\%$, only a marginal $1.0\%$ drop) while achieving the same high robustness gains ($81.0\%$ Consistency) as the naive approach. This demonstrates that our method successfully decouples robustness from in-domain degradation.

\subsection{Stability Analysis: Robustness to Data Scaling}

\begin{figure*}[t]
    \centering
    \includegraphics[width=0.95\textwidth]{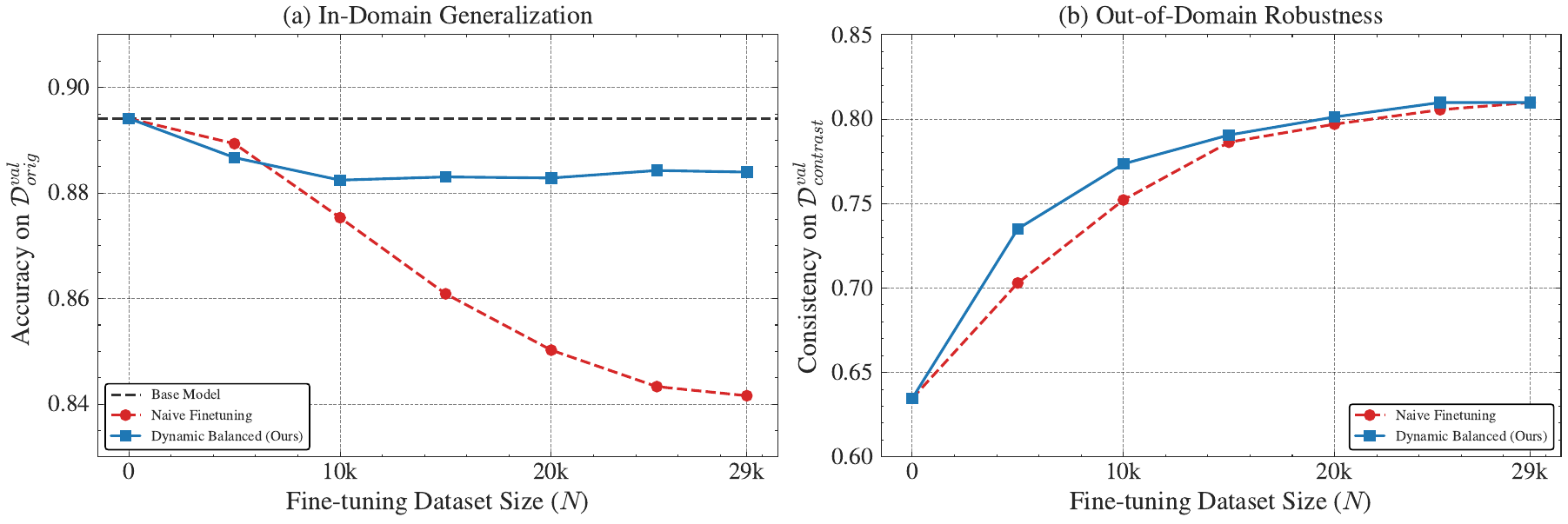} 
    
    \caption{\textbf{Impact of fine-tuning dataset size on model performance.} 
    \textbf{(a)} Shows generalization on the original SNLI validation set; Naive Finetuning (red) suffers from catastrophic forgetting as data scales, whereas our Dynamic Balanced strategy (blue) maintains performance near the Base Model. 
    \textbf{(b)} Shows robustness on the synthetic contrast set; both methods achieve similar gains in consistency, demonstrating that our strategy prevents forgetting without compromising the learning of new boundaries.}
    \label{fig:scaling_results}
\end{figure*}

    To validate the stability of our proposed method, we analyzed the performance trajectories as a function of the fine-tuning dataset size $N$), ranging from $0$ to $29\text{k}$ examples. Figure~\ref{fig:scaling_results} illustrates the divergent behaviors of the two strategies.\paragraph{Impact on Generalization (Fig.~\ref{fig:scaling_results}a).}The naive finetuning strategy (red line) exhibits a monotonic degradation in original accuracy as the volume of synthetic data increases. The accuracy drops from the baseline of $89.4\%$ down to $84.1\%$ when using the full contrast set. This provides strong empirical evidence that catastrophic forgetting is directly correlated with the magnitude of the distribution shift—the more artifact-free data the model sees, the more it overwrites its original priors. In sharp contrast, our Dynamic Balanced strategy (blue line) remains effectively pinned near the baseline performance ($\approx 88.4\%$) regardless of the dataset size. The flat trajectory confirms that mixing the original distribution prevents the model weights from drifting too far from the general NLI manifold.\paragraph{Scaling of Robustness (Fig.~\ref{fig:scaling_results}b).}Crucially, Figure~\ref{fig:scaling_results}b demonstrates that this stability does not come at the cost of robustness. Both strategies show a nearly identical logarithmic growth in Consistency Score, saturating around $81\%$. This implies that the model's ability to learn the new decision boundaries is driven primarily by the quality of the contrast examples, while the retention of old knowledge is determined by the training curriculum. Our method achieves the ``best of both worlds'': maximum robustness with minimum forgetting.

\section{Conclusion}

In this work, we addressed the critical reliance of NLI models on annotation artifacts, a pervasive issue in large-scale crowd-sourced benchmarks like \textbf{SNLI}. To tackle this, we proposed an end-to-end automated framework for robustification. First, we introduced \textbf{LF-LMI}, a metric that effectively isolates semantic artifacts from frequency noise, allowing us to systematically diagnose spurious correlations within the training data. Leveraging this diagnostic, we developed a scalable pipeline to synthesize high-quality contrast sets via Large Language Models, validated by a rigorous multi-judge consensus protocol.

Furthermore, we demonstrated that standard fine-tuning on such hard examples leads to catastrophic forgetting of the original data distribution. To solve this, we proposed \textbf{Dynamic Balanced Sampling}, a training curriculum that rotates the original data in every epoch while anchoring the contrast set. Our experiments confirm that this strategy significantly improves robustness ($+17.5\%$ consistency) without compromising in-domain accuracy ($88.4\%$). This work provides a reproducible blueprint for mitigating dataset-specific biases, decoupling the need for expensive human annotation from the pursuit of robust, reasoning-capable models.


\begin{thebibliography}{9}
\expandafter\ifx\csname natexlab\endcsname\relax\def\natexlab#1{#1}\fi

\bibitem[{Bowman et~al.(2015)Bowman, Angeli, Potts, and Manning}]{bowman2015large}
Samuel~R. Bowman, Gabor Angeli, Christopher Potts, and Christopher~D. Manning. 2015.
\newblock \href {https://doi.org/10.18653/v1/D15-1075} {A large annotated corpus for learning natural language inference}.
\newblock In \emph{Proceedings of the 2015 Conference on Empirical Methods in Natural Language Processing}, pages 632--642, Lisbon, Portugal. Association for Computational Linguistics.

\bibitem[{Clark et~al.(2020)Clark, Luong, Le, and Manning}]{clark2020electra}
Kevin Clark, Minh-Thang Luong, Quoc~V. Le, and Christopher~D. Manning. 2020.
\newblock \href {https://openreview.net/pdf?id=r1xMH1BtvB} {Electra: Pre-training text encoders as discriminators rather than generators}.
\newblock In \emph{Proceedings of the 8th International Conference on Learning Representations (ICLR)}.

\bibitem[{Evert(2005)}]{evert2005}
Stefan Evert. 2005.
\newblock \emph{The Statistics of Word Co-occurrences: Word Pairs and Collocations}.
\newblock Dissertation, University of Stuttgart.

\bibitem[{Gardner et~al.(2020)Gardner, Artzi, Basmov, Berant, Bogin, Chen, Dasigi, Dua, Elazar, Gottumukkala, Gupta, Hajishirzi, Ilharco, Khashabi, Lin, Liu, Liu, Mulcaire, Ning, Singh, Smith, Subramanian, Tsarfaty, Wallace, Zhang, and Zhou}]{gardner2020}
Matt Gardner, Yoav Artzi, Victoria Basmov, Jonathan Berant, Ben Bogin, Sihao Chen, Pradeep Dasigi, Dheeru Dua, Yanai Elazar, Ananth Gottumukkala, Nitish Gupta, Hanna Hajishirzi, Gabriel Ilharco, Daniel Khashabi, Kevin Lin, Jiangming Liu, Nelson~F. Liu, Phoebe Mulcaire, Qiang Ning, Sameer Singh, Noah~A. Smith, Sanjay Subramanian, Reut Tsarfaty, Eric Wallace, Ally Zhang, and Ben Zhou. 2020.
\newblock \href {https://doi.org/10.18653/v1/2020.findings-emnlp.117} {Evaluating models’ local decision boundaries via contrast sets}.
\newblock In \emph{Findings of the Association for Computational Linguistics: EMNLP 2020}, pages 1307--1323, Online. Association for Computational Linguistics.

\bibitem[{Gururangan et~al.(2018)Gururangan, Swayamdipta, Levy, Schwartz, Bowman, and Smith}]{gururangan2018}
Suchin Gururangan, Swabha Swayamdipta, Omer Levy, Roy Schwartz, Samuel~R. Bowman, and Noah~A. Smith. 2018.
\newblock \href {https://doi.org/10.18653/v1/N18-2017} {Annotation artifacts in natural language inference data}.
\newblock In \emph{Proceedings of the 2018 Conference of the North American Chapter of the Association for Computational Linguistics: Human Language Technologies, Volume 2 (Short Papers)}, pages 107--112, New Orleans, Louisiana. Association for Computational Linguistics.

\bibitem[{Luo et~al.(2023)Luo, Yang, Meng, Li, Zhou, and Zhang}]{luo2023empirical}
Yun Luo, Zhen Yang, Fandong Meng, Yafu Li, Jie Zhou, and Yue Zhang. 2023.
\newblock \href {https://arxiv.org/abs/2308.08747} {An empirical study of catastrophic forgetting in large language models during continual fine-tuning}.
\newblock \emph{arXiv preprint arXiv:2308.08747}.

\bibitem[{Poliak et~al.(2018)Poliak, Naradowsky, Haldar, Rudinger, and Durme}]{poliak2018}
Adam Poliak, Jason Naradowsky, Aparajita Haldar, Rachel Rudinger, and Benjamin~Van Durme. 2018.
\newblock \href {https://doi.org/10.18653/v1/S18-2023} {Hypothesis only baselines in natural language inference}.
\newblock In \emph{Proceedings of the 7th Joint Conference on Lexical and Computational Semantics (*SEM 2018)}, pages 180--191, New Orleans, Louisiana. Association for Computational Linguistics.

\bibitem[{Schuster et~al.(2019)Schuster, Shah, Yeo, Filizzola, Santus, and Barzilay}]{schuster2019}
Tal Schuster, Darsh~J. Shah, Yun Jie~Serene Yeo, Daniel Filizzola, Enrico Santus, and Regina Barzilay. 2019.
\newblock \href {https://doi.org/10.18653/v1/D19-1341} {Towards debiasing fact verification models}.
\newblock In \emph{Proceedings of the 2019 Conference on Empirical Methods in Natural Language Processing and the 9th International Joint Conference on Natural Language Processing (EMNLP-IJCNLP)}, pages 3419--3425, Hong Kong, China. Association for Computational Linguistics.

\bibitem[{Wu et~al.(2021)Wu, Ribeiro, Heer, and Weld}]{wu2021polyjuice}
Tongshuang Wu, Marco~Tulio Ribeiro, Jeffrey Heer, and Daniel~S. Weld. 2021.
\newblock \href {https://doi.org/10.18653/v1/2021.acl-long.521} {Polyjuice: Generating counterfactual explanations for explaining and debugging nlp models}.
\newblock In \emph{Proceedings of the 59th Annual Meeting of the Association for Computational Linguistics and the 11th International Joint Conference on Natural Language Processing (Volume 1: Long Papers)}, pages 6674--6690, Online. Association for Computational Linguistics.

\end{thebibliography}

\appendix

\section{Appendices}
\label{sec:appendix}

\subsection{Additional LF-LMI Artifact Analysis}
\label{app:lmi_tables}

In Section 3.2, we presented the analysis for the \textsc{Contradiction} label. Here, we extend this validation to the \textsc{Entailment} and \textsc{Neutral} classes. Table~\ref{tab:app_lmi_full} provides the side-by-side comparison.

For \textsc{Entailment}, LF-LMI prioritizes high-precision generalizations (e.g., \textit{a human}, \textit{is outdoors}) with conditional probabilities $P(l|w) > 0.80$, filtering out the noisy function words (\textit{a man}, \textit{there is}) that dominate standard LMI. Similarly, for \textsc{Neutral}, LF-LMI effectively captures specific modifiers (\textit{tall human}, \textit{sad}) and temporal markers (\textit{waiting for}) that reflect the ``inference/hallucination'' strategy, whereas standard LMI is cluttered with less discriminative prepositions like \textit{for a}.

\begin{table*}[h]
    \centering
    \small
    \renewcommand{\arraystretch}{1.1}
    \setlength{\tabcolsep}{10pt}
    
    \begin{tabular}{cc}
        \multicolumn{2}{c}{\textbf{(a) ENTAILMENT LABEL}} \\
        \midrule
        \begin{tabular}{lrrr}
            \multicolumn{4}{c}{\textit{LF-LMI (Ours)}} \\
            \textbf{Bigram} & \textbf{Val} & \textbf{Freq} & \textbf{$P$} \\
            \midrule
            \textit{a human}      & 8.17 & 1236 & 0.93 \\
            \textit{is outside}   & 8.16 & 4652 & 0.79 \\
            \textit{are outside}  & 8.12 & 3754 & 0.81 \\
            \textit{is outdoors}  & 7.97 & 1554 & 0.88 \\
            \textit{are outdoors} & 7.81 & 1214 & 0.89 \\
            \textit{this picture} & 7.73 & 817 & 0.94 \\
            \textit{is near}      & 7.28 & 1261 & 0.83 \\
            \textit{an animal}    & 7.17 & 1185 & 0.83 \\
            \textit{at least}     & 6.80 & 386 & 0.93 \\
            \textit{an instrument}& 6.47 & 803 & 0.80 \\
        \end{tabular} 
        & 
        \begin{tabular}{lrrr}
            \multicolumn{4}{c}{\textit{Standard LMI}} \\
            \textbf{Bigram} & \textbf{Val} & \textbf{Freq} & \textbf{$P$} \\
            \midrule
            \textit{a man}        & 6066 & 65744 & 0.37 \\
            \textit{a person}     & 5536 & 13034 & 0.60 \\
            \textit{there are}    & 5450 & 10653 & 0.65 \\
            \textit{there is}     & 4813 & 12481 & 0.57 \\
            \textit{people are}   & 4039 & 32517 & 0.40 \\
            \textit{is outside}   & 3655 & 4652 & 0.79 \\
            \textit{are outside}  & 3063 & 3754 & 0.81 \\
            \textit{is a}         & 2750 & 14190 & 0.45 \\
            \textit{two people}   & 2466 & 8124 & 0.52 \\
            \textit{person is}    & 2231 & 9470 & 0.48 \\
        \end{tabular} \\
    \end{tabular}
    
    \vspace{0.5cm} 
    
    \begin{tabular}{cc}
        \multicolumn{2}{c}{\textbf{(b) NEUTRAL LABEL}} \\
        \midrule
        \begin{tabular}{lrrr}
            \multicolumn{4}{c}{\textit{LF-LMI (Ours)}} \\
            \textbf{Bigram} & \textbf{Val} & \textbf{Freq} & \textbf{$P$} \\
            \midrule
            \textit{tall human}   & 6.20 & 587 & 1.00 \\
            \textit{a tall}       & 6.09 & 1199 & 0.90 \\
            \textit{a sad}        & 5.81 & 671 & 0.93 \\
            \textit{for his}      & 5.76 & 1764 & 0.83 \\
            \textit{tall humans}  & 5.72 & 364 & 0.99 \\
            \textit{sad man}      & 5.45 & 323 & 0.97 \\
            \textit{practicing}   & 5.43 & 381 & 0.95 \\
            \textit{tall person}  & 5.36 & 285 & 0.98 \\
            \textit{first time}   & 5.29 & 458 & 0.91 \\
            \textit{on vacation}  & 5.27 & 403 & 0.92 \\
        \end{tabular} 
        & 
        \begin{tabular}{lrrr}
            \multicolumn{4}{c}{\textit{Standard LMI}} \\
            \textbf{Bigram} & \textbf{Val} & \textbf{Freq} & \textbf{$P$} \\
            \midrule
            \textit{for a}        & 3041 & 7685 & 0.68 \\
            \textit{for the}      & 1952 & 4648 & 0.69 \\
            \textit{trying to}    & 1684 & 3591 & 0.72 \\
            \textit{waiting for}  & 1530 & 2994 & 0.75 \\
            \textit{for his}      & 1156 & 1764 & 0.83 \\
            \textit{about to}     & 1135 & 2506 & 0.71 \\
            \textit{going to}     & 1113 & 2047 & 0.77 \\
            \textit{is trying}    & 1081 & 2167 & 0.74 \\
            \textit{to get}       & 941 & 1915 & 0.74 \\
            \textit{to the}       & 865 & 4895 & 0.53 \\
        \end{tabular} \\
    \end{tabular}
    
    \caption{Top spurious bigrams retrieved by LF-LMI vs. Standard LMI for the \textsc{Entailment} (top) and \textsc{Neutral} (bottom) labels. In both cases, LF-LMI successfully dampens high-frequency noise (e.g., \textit{a man}, \textit{for a}), prioritizing n-grams with higher conditional probabilities ($P$) that represent true semantic artifacts.}
    \label{tab:app_lmi_full}
\end{table*}

\subsection{Prompt Engineering}
\label{app:prompts}

To ensure reproducibility, we provide the exact prompts used. 

\paragraph{Counterfactual Synthesis Prompt}
This prompt enforces minimal edits on the premise while anchoring the hypothesis.

\begin{lstlisting}
Act as an expert NLI (Natural Language Inference) dataset augmentor.
Your goal is to generate a "Contrast Set" by modifying the PREMISE.

TASK:
You are provided with an (Original Premise, Hypothesis) pair.
You must generate a **NEW PREMISE** by MINIMALLY editing the Original Premise so that the relationship with the (unchanged) Hypothesis becomes: **{TARGET_LABEL}**.

INPUT DATA:
- Original Premise: "{PREMISE}"
- Hypothesis (DO NOT CHANGE): "{HYPOTHESIS}"
- Target Label: {TARGET_LABEL}

GUIDELINES:
1. **Minimal Edit:** Change as few words as possible in the Original Premise (e.g., change the subject, the action, or the setting). Do not rewrite the entire sentence unless necessary.
2. **Single Scene Logic:** Assume the text describes a single visual scene.
   - If Target is **Contradiction**: The New Premise must describe a scene that makes the Hypothesis physically impossible.
   - If Target is **Entailment**: The New Premise must explicitly confirm the details in the Hypothesis, but not be exactly equal to it.
   - If Target is **Neutral**: The New Premise should describe a relevant scene but leave the specific details of the Hypothesis unknown.
3. **Consistency:** The New Premise must be grammatically correct and natural.

OUTPUT FORMAT:
Return ONLY the text of the New Premise. Do not add labels or explanations.

New Premise:
\end{lstlisting}

\paragraph{LLM-as-a-Judge Validation Prompt}
This prompt is utilized by two independent judges. Only generations with unanimous approval are retained.

\begin{lstlisting}
Act as an NLI Quality Assurance validator. Your task is to accept or reject a generated "Contrast Set" example.

INPUTS:
- Original Premise: "{PREMISE}"
- Hypothesis: "{HYPOTHESIS}"
- New Premise: "{NEW_PREMISE}"
- Target Label: {TARGET_LABEL}

CORE ASSUMPTION:
Treat the New Premise as an **EXHAUSTIVE** description of the entire scene/photo. Assume that any person, object, or animal NOT mentioned in the New Premise does NOT exist in the scene.

CRITERIA FOR APPROVAL (Must meet BOTH):
1. Minimal Perturbation: The New Premise must be a modified version of the Original Premise (preserving structure and context). It must NOT be a completely unrelated sentence.
2. Label Accuracy: Based on the exhaustive Core Assumption, the logical relationship between the "New Premise" and the "Hypothesis" must strictly match the "{TARGET_LABEL}".
3. Naturalness: The New Premise must be a natural sentence, not forced or unnatural.

OUTPUT FORMAT:
Return ONLY a string (no markdown, no conversational text):
<valid>|<short_concise_reasoning>
Where valid is true if the new premise is a valid contrast set, and false otherwise.
\end{lstlisting}

\end{document}